 \documentclass[pmlr,twocolumn]{jmlr} 

\usepackage{booktabs}
\usepackage{amsmath}
\usepackage{graphicx}
\usepackage{adjustbox}
\usepackage{float}
\usepackage[load-configurations=version-1]{siunitx} 

\theorembodyfont{\upshape}
\theoremheaderfont{\scshape}
\theorempostheader{:}
\theoremsep{\newline}

\jmlrvolume{ML4H Extended Abstract Arxiv Index}
\jmlryear{2020}
\jmlrsubmitted{2020}
\jmlrpublished{}
\jmlrworkshop{Machine Learning for Health (ML4H) 2020}

\title[Domain Generalization for Ultrasound Segmentation Using Transfer Learning]{A Study of Domain Generalization on Ultrasound-based Multi-Class Segmentation of Arteries, Veins, Ligaments, and Nerves Using Transfer Learning}
\author{%
\Name{Edward Chen, Tejas Sudharshan Mathai, Vinit Sarode, Howie Choset, and John Galeotti} \Email{edwardc2@cs.cmu.edu}\\
\addr Carnegie Mellon University, Pittsburgh, Pennsylvania, 15289, USA
}

\begin{document}

\maketitle

\begin{abstract}
Identifying landmarks in the femoral area is crucial for ultrasound (US) -based robot-guided catheter insertion, and their presentation varies when imaged with different scanners. As such, the performance of past deep learning-based approaches is also narrowly limited to the training data distribution; this can be circumvented by fine-tuning all or part of the model, yet the effects of fine-tuning are seldom discussed. In this work, we study the US-based segmentation of multiple classes through transfer learning by fine-tuning different contiguous blocks within the model, and evaluating on a gamut of US data from different scanners and settings. We propose a simple method for predicting generalization on unseen datasets and observe statistically significant differences between the fine-tuning methods while working towards domain generalization.

\end{abstract}

\section{Introduction}
\label{sec:intro}


In the case of major internal hemorrhaging, real-time ultrasound (US) imaging can guide the robotic insertion of a vascular catheter for Resuscitative Endovascular Balloon Occlusion of the Aorta (REBOA) via the femoral artery to prevent the patient from bleeding to death. Automatically segmenting femoral area landmarks will be crucial to the optimal catheter placement in time-sensitive situations. To this end, the developed technology has to be robust to variations in anatomy, scanner settings, external artifacts (in traumatic injury scenarios), probe positioning, etc. However, medical imaging datasets are often limited in quantity and span a restricted distribution over the data space \citep{parker-system-medical-data}. Deep learning models trained on such data perform poorly when tested on data from different anatomic areas or scanner settings \citep{mani-coverage-testing-19}, thereby limiting their real-world usage. For instance, \citep{chen-nature-2020} illustrate the first-attempt success rate for ultrasound-guided needle insertions dropping by $\sim$30\% on datasets of different anatomy and settings, leaving room for serious consequences to the patient \citep{dudeck-ultrasound-complications-2004}.


The present standard for improving robustness in the medical domain is to use an existing architecture trained on natural images, such as ImageNet \citep{russakovsky-imagenet, geirhos-imagenet-texture-2019}, and then fine-tune on the medical images \citep{raghu-transfusion, iglovikov-ternausnet-2018}. However, little work has been done to illuminate the generalization ability of models to medical images using transfer learning, and to understand segmentation models commonly used in medical robotics \citep{pakhomav-robotic-surgery-2019}, such as the U-Net \citep{ronneberger-unet}. Raghu, et. al. \citep{raghu-transfusion} performed an in-depth study of transfer learning for classification of medical images, albeit starting from natural images. Amiri, et. al. \citep{amiri-fine-tuning-2019} studied UNet-based fine-tuning, but on a single domain.


We aim to expand the current understanding of domain generalization and provide actionable insights for enhanced robustness within the context of ultrasound-based multi-class segmentation using transfer learning. Here, we consider the practical case where the data is gathered in a sequential manner, specifically when the previously trained data is unavailable due to privacy restrictions \citep{patel-medical-imaging-sharing-2019}. In reality, deep learning models for medical devices will also often have been trained on some subset of medical imaging domains, for which labelled data was available. We further design the experiments in a way to more explicitly control for certain training domains, attempting to generalize training across more of the real-world space of unexpected images. 

From our experiments, we reveal the following insights: (1) As consecutive blocks on both the encoder and decoder side are individually fine-tuned, the out-of-training-domain (\emph{OOTD}) performance generally increases. The \emph{OOTD} data are different from the pre-training data (\emph{pt-data}) and fine-tuning data (\emph{ft-data}). (2) Having a smaller number of classes in the \emph{pt-data} may hamper the final performance on the \emph{ft-data}, but not of that for the \emph{OOTD} data, and (3) There is a statistically significant difference between fine-tuning the encoder and decoder in terms of performance on \emph{OOTD} data. We then take into account such observations and propose selecting the \emph{ft-data} performance as a proxy for OOTD performance when selecting the best fine-tuning method to use.


\section{Materials And Methods}

\noindent
\textbf{Scanners}: 3 different ultrasound scanners with diverse scan settings (e.g. gain values) were used: (1) a portable scanner, (2) a high-frequency ultrasound (\emph{HFUS}) scanner, and (3) an ultra high-frequency ultrasound (\emph{UHFUS}) scanner. More details in \appendixref{apd:data}. 

\noindent
\textbf{Human Data}: The UHFUS scanner imaged arteries and veins in human subjects \citep{mathai-fast-vessel-segmentation-2018}, and a single class label was assigned to them by an expert. We refer to this data as \emph{human-single50} (\emph{h50}). 

\noindent
\textbf{Phantom Data}: 3 categories of sequences were acquired from a phantom: (1) \emph{phantom1-multi12} (\emph{ph1-12}), (2) \emph{phantom1-multi22} (\emph{ph1-22}), and (3) \emph{phantom2-multi12} (\emph{ph2-12}). The prefix ``phantom1'' represents image sequences collected from the left side of the phantom, while prefix ``phantom2'' represents data collected from the right side of the phantom, which also contained different anatomy (muscles, liver, etc.). 4 classes were labelled in each phantom dataset: arteries, veins, ligaments, and nerves. 

\noindent
\textbf{Pig Data}: Data was gathered from a living pig using the portable scanner, and the arteries and veins were labelled as 2 classes. We refer to this as \emph{pig-multi12} (\emph{p12}). Each numerical suffix represents the frequency with which it was collected with. More details in \appendixref{apd:data}.

\begin{figure}[ht]
\includegraphics[width=.08\textwidth]{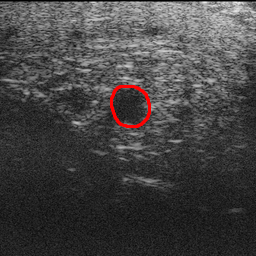}\hfill
\includegraphics[width=.08\textwidth]{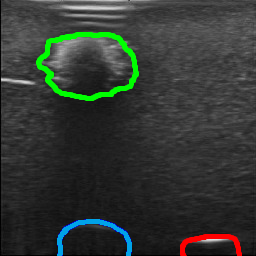}\hfill
\includegraphics[width=.08\textwidth]{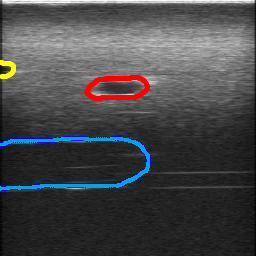}\hfill
\includegraphics[width=.08\textwidth]{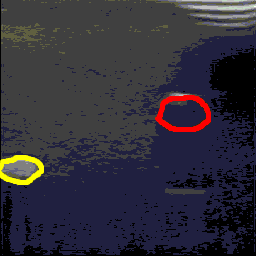}\hfill
\includegraphics[width=.08\textwidth]{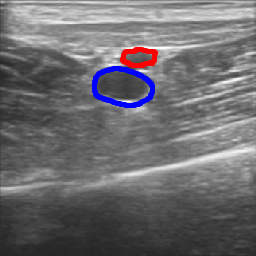}\hfill
\caption{Sample images of (from the left): \emph{h50}, \emph{ph1-12}, \emph{ph1-22}, \emph{ph2-12}, \emph{p12}}
    \label{fig:all-datasets}
\end{figure}



\noindent
\textbf{Transfer Learning}: We evaluated segmentation performance on a U-Net model consisting of 5 encoder blocks (including the bottleneck layer as the 5th encoder) and 4 decoder blocks. We first transferred previously learned weights and then fine-tuned different models that spanned various \textbf{contiguous} blocks of the architecture. The encoder blocks were numbered from 1 to 5, starting from the input layer and ending with the bottleneck layer. The numbering for the encoder side is cumulative, e.g. "Encoder 5" refers to all 5 of the blocks leading up to the bottleneck layer. The decoder blocks were numbered from 4 (just after the bottleneck) to 1 for the output layer. "Decoder 1" in this case refers to the 4 blocks of the decoder up to the output layer. \figureref{fig:u-net-naming} illustrates this naming convention.  
  
\begin{figure}[htbp]
\floatconts
    {fig:u-net-naming}
    {\caption{Our U-Net naming convention}}
    {\includegraphics[width=.24\textwidth]{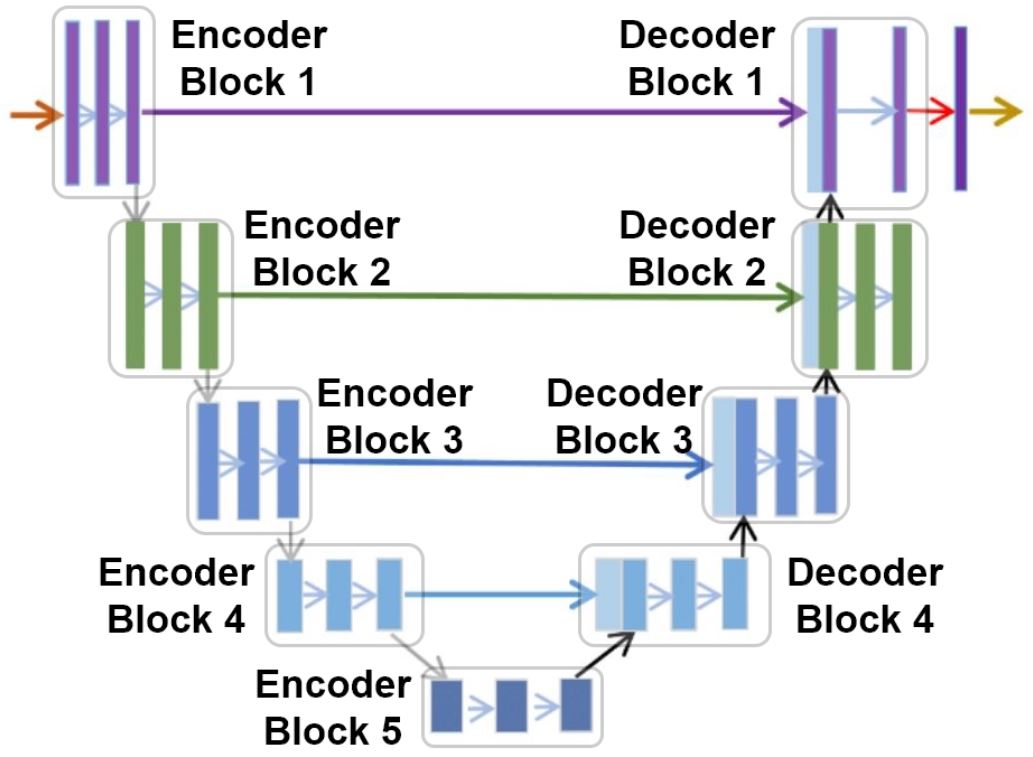}}
\end{figure}

\section{Experiments and Results} \label{sec:experiments-results}
\textbf{Design}: To mimic clinical scenarios, we tested the various models on 2 variations of data: (1) anatomy and (2) imaging parameters. To illustrate the changes in the data distribution over anatomy (based on class count) and imaging settings, we compared fine-tuning procedures across \emph{h50} and \emph{ph1-12}. To compare against changes solely in imaging parameters, we used \emph{ph1-22} and \emph{ph1-12}. We conducted four experiments (paired as \{\emph{pt-data}, \emph{ft-data}\}): (1) \{\emph{ph1-12}, \emph{h50}\} (2) \{\emph{h50}, \emph{ph1-12}\} (3) \{\emph{ph1-12},  \emph{ph1-22}\} and (4) \{\emph{ph1-22}, \emph{ph1-12}\}. In addition to fine-tuning the entire model, each pair was evaluated across all permutations of contiguous blocks on the encoder and decoder side. The \emph{OOTD} datasets comprised the 3 left-out datasets that were not being directly compared against. We measured the \emph{OOTD} performance by taking the arithmetic average across the unseen datasets' scores. To replicate the lack of access to the \emph{pt-data} in our scenario, we did not provide scores for that category in each experiment. We evaluate results using the Dice Similarity Coefficient (DSC). Full numerical results are seen in \appendixref{apd:results} from Tables \ref{table:transfer-results-details-1} and \ref{table:transfer-results-details-2} with training details described in \appendixref{apd:training}.

\noindent
\textbf{Discussion}: 
For all 4 of the experiments, we noted a general trend: (1) as a larger number of the encoder/decoder blocks are fine-tuned, the model performed equally well or better on data from both its \emph{ft-data} domain and \emph{OOTD} data. We visualize this pattern in Figures \ref{fig:encoder-blocks-ootd-plot} and \ref{fig:decoder-blocks-ootd-plot}. \appendixref{apd:results} describes significance testing. We note that the one exception to this is the decoder branch of experiment 2 (\figureref{fig:decoder-blocks-ootd-plot}), which leads to our next observation: (2) for models in experiment 2, we noticed that it was ``difficult'' for the batch normalization statistics to converge during the fine-tuning process. We believe this to be due to the fewer number of classes in the \emph{pt-data} domain, leading to a more restrictive feature representation. On the contrary, the opposite is true for experiment 1, which has contiguous block-wise performances close to, and even surpassing, that of the full-model fine-tuning procedure. We attempt to further understand this class count-related effect together with another observation later in this section. 

\begin{figure}[t]
\floatconts
    {fig:encoder-blocks-ootd-plot}
    {\caption{Best-fit lines showing positive relationship between longer encoder subsequences and \emph{OOTD} scores for each of the 4 experiments}}
    {\includegraphics[width=.44\textwidth]{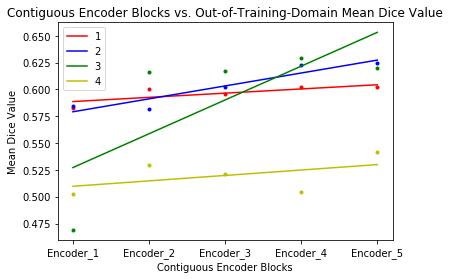}}
\end{figure}

We additionally notice that (3) fine-tuning contiguous encoder blocks produced better \emph{OOTD} performance than those with decoder blocks, which can also be seen in Tables \ref{table:transfer-results-details-1} and \ref{table:transfer-results-details-2} (\appendixref{apd:results}). This may be because fine-tuning blocks on the encoder side leads to a more diverse latent feature representation while retaining the localization information on the decoder side. We conducted Wilcoxon tests on the paired encoder-decoder differences for each of the 4 experiments. All had statistically significant greater \emph{OOTD} performance from the encoder branch, except experiment 1, which may have had the decoder benefit from the greater number of classes in \emph{pt-data}.


Considering the above, we propose to use each fine-tuning method's \emph{ft-data} performance as a representation of its \emph{OOTD} score. Our proposed method is to select the contiguous-block-wise fine-tuned model with the highest \emph{ft-data} score. Note that the full-model fine-tuning always produced worse \emph{OOTD} scores in our case. We can use such a method to predict a fine-tuned model that might have the best domain generalization capabilities while simultaneously selecting a fine-tuned model that performs well in its direct target task. We compare the \emph{OOTD} performance of our method's model choices with that of the traditional full-model fine-tuning method in Table \ref{proposed-method-table}.

\begin{figure}[htbp]
\floatconts
    {fig:decoder-blocks-ootd-plot}
    {\caption{Best-fit lines showing positive relationship between longer decoder subsequences and \emph{OOTD} scores, except for case 2 which may be affected by class count}}
    {\includegraphics[width=.44\textwidth]{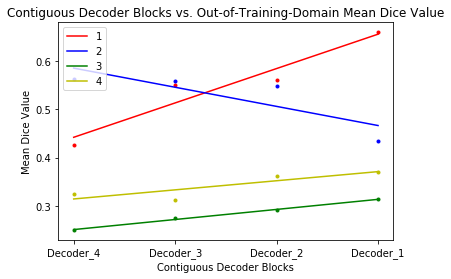}}
\end{figure}

In all cases, our method surpasses the \emph{OOTD} performance with the full-model fine-tuning method by a large margin; the same occurs against those of decoder-only methods, which are also commonly used. Of note are experiments 2 and 3, where the chosen method's \emph{ft-data} performance is lower than that of the full-model method - despite a higher \emph{OOTD} score. We observe in both of these cases that the \emph{ft-data} is objectively and quantitatively more difficult than the \emph{pt-data} (which may also explain (2) above). To quantify their difficulties, we use the autoencoder reconstruction error \citep{xia-unsupervised-discriminative-2015}. We summarize that our proposed method results in the near-optimal \emph{OOTD} performance in \textbf{4/4} of the cases. Based on this result, we suggest that our use of an autoencoder might be an effective general approach to address \emph{a priori} the trade-off between \emph{ft-data} and \emph{OOTD} performance. Further details are provided in \appendixref{apd:results}.

\begin{table}[h]
    \caption[]{Proposed Method for Enhanced Generalization using \emph{OOTD} Scores}
    \label{proposed-method-table}
    \begin{center}
    \setlength{\tabcolsep}{2pt}
    \begin{tabular}{|c||c||c|}
    \hline
    No. & Ours (Method) & Full \\
    \hline
    1 & \textbf{.656 $\pm$ .0498} (d-1) & .612 $\pm$ .0383 \\
    \hline
    2 & \textbf{.625 $\pm$ .0346} (e-5) & .328 $\pm$ .0919 \\
    \hline
    3 & \textbf{.620 $\pm$ .0820} (e-5) & .393 $\pm$ .1175 \\
    \hline
    4 & \textbf{.541 $\pm$ .1702} (e-5) & .403 $\pm$ .1776 \\
    \hline
    \end{tabular}
    \end{center}
\end{table}

\bibliography{jmlr-sample}

\begin{thebibliography}{17}
\providecommand{\natexlab}[1]{#1}
\providecommand{\url}[1]{\texttt{#1}}
\expandafter\ifx\csname urlstyle\endcsname\relax
  \providecommand{\doi}[1]{doi: #1}\else
  \providecommand{\doi}{doi: \begingroup \urlstyle{rm}\Url}\fi

\bibitem[Abadi et~al.(2016)Abadi, Barham, Chen, Chen, Davis, Dean, Devin,
  Ghemawat, Irving, Isard, Kudlur, Levenberg, Monga, Moore, Murray, Steiner,
  Tucker, Vasudevan, Warden, Wicke, Yu, and Zheng]{martin-tensorflow-2016}
M.~Abadi, P.~Barham, J.~Chen, Z.~Chen, A.~Davis, J.~Dean, M.~Devin,
  S.~Ghemawat, G.~Irving, M.~Isard, M.~Kudlur, J.~Levenberg, R.~Monga,
  S.~Moore, D.~Murray, B.~Steiner, P.~Tucker, V.~Vasudevan, P.~Warden,
  M.~Wicke, Y.~Yu, and X.~Zheng.
\newblock Tensorflow: A system for large-scale machine learning.
\newblock In \emph{USENIX Symposium on Operating Systems Design and
  Implementation}, 2016.

\bibitem[Amiri et~al.(2019)Amiri, Brooks, and Rivaz]{amiri-fine-tuning-2019}
M.~Amiri, R.~Brooks, and H.~Rivaz.
\newblock Fine tuning u-net for ultrasound image segmentation: which layers?
\newblock In \emph{International conference on medical image computing and
  computer-assisted intervention}, 2019.

\bibitem[Chen et~al.(2020)Chen, Balter, Maguire, and Yarmush]{chen-nature-2020}
A.~Chen, M.~Balter, T.~Maguire, and M.~Yarmush.
\newblock Deep learning robotic guidance for autonomous vascular access.
\newblock \emph{Nature Machine Intelligence}, 2020.

\bibitem[Dudeck et~al.(2004)Dudeck, Teichgraeber, Podrabsky, Haenninen,
  Soerensen, and Ricke]{dudeck-ultrasound-complications-2004}
O.~Dudeck, U.~Teichgraeber, P.~Podrabsky, E.~Haenninen, R.~Soerensen, and
  J.~Ricke.
\newblock A randomized trial assessing the value of ultrasound-guided puncture
  of the femoral artery for interventional investigations.
\newblock \emph{The International Journal of Cardiovascular Imaging}, 2004.

\bibitem[Geirhos et~al.(2019)Geirhos, Rubish, Michaelis, Bethge, Wichmann, and
  Brendel]{geirhos-imagenet-texture-2019}
R.~Geirhos, P.~Rubish, C.~Michaelis, M.~Bethge, F.~Wichmann, and W.~Brendel.
\newblock Imagenet-trained cnns are biased towards texture: increasing shape
  bias improves accuracy and robustness.
\newblock In \emph{International Conference on Learning Representations}, 2019.

\bibitem[Iglovikov and Shvletz(2018)]{iglovikov-ternausnet-2018}
V.~Iglovikov and A.~Shvletz.
\newblock Ternausnet: U-net with vgg11 encoder pre-trained on imagenet for
  image segmentation.
\newblock \emph{arXiV}, 2018.

\bibitem[Kingma and Ba(2014)]{kingma-adam-2014}
D.~Kingma and J.~Ba.
\newblock Adam: A method for stochastic optimization.
\newblock 2014.

\bibitem[Mani et~al.(2019)Mani, Sankaran, Tamilselvam, and
  Sethi]{mani-coverage-testing-19}
S.~Mani, A.~Sankaran, S.~Tamilselvam, and A.~Sethi.
\newblock Coverage testing of deep learning models using dataset
  characterization.
\newblock \emph{arXiV}, 2019.

\bibitem[Mathai et~al.(2018)Mathai, Jin, Gorantla, and
  Galeotti]{mathai-fast-vessel-segmentation-2018}
T.S. Mathai, L.~Jin, V.~Gorantla, and J.~Galeotti.
\newblock Fast vessel segmentation and tracking in ultra high-frequency
  ultrasound images.
\newblock In \emph{International Conference on Medical Image Computing and
  Computer-Assisted Intervention}, 2018.

\bibitem[Mathai et~al.(2019)Mathai, Gorantla, and Galeotti]{mathai-uhfus-2019}
T.S. Mathai, V.~Gorantla, and J.~Galeotti.
\newblock Segmentation of vessels in ultra high frequency ultrasound sequences
  using contextual memory.
\newblock In \emph{International Conference on medical image computing and
  computer-assisted intervention}, 2019.

\bibitem[Pakhomav et~al.(2019)Pakhomav, Premachandran, Allan, Azizian, and
  Navab]{pakhomav-robotic-surgery-2019}
D.~Pakhomav, V.~Premachandran, M.~Allan, M.~Azizian, and N.~Navab.
\newblock Deep residual learning for instrument segmentation in robotic
  surgery.
\newblock In \emph{International Workshop on Machine Learning in Medical
  Imaging}, 2019.

\bibitem[Parker(2006)]{parker-system-medical-data}
J.~Parker.
\newblock System and method for managing medical data, 2006.

\bibitem[Patel(2019)]{patel-medical-imaging-sharing-2019}
V.~Patel.
\newblock A framework for secure and decentralized sharing of medical imaging
  data via blockchain consensus.
\newblock \emph{Health Informatics Journal}, 2019.

\bibitem[Raghu et~al.(2019)Raghu, Zhang, Kleinberg, and
  Bengio]{raghu-transfusion}
M.~Raghu, C.~Zhang, J.~Kleinberg, and S.~Bengio.
\newblock Transfusion: Understanding transfer learning for medical imaging.
\newblock In \emph{Advances in Neural Information Processing Systems}, 2019.

\bibitem[Ronneberger et~al.(2015)Ronneberger, Fischer, and
  Brox]{ronneberger-unet}
O.~Ronneberger, P.~Fischer, and T.~Brox.
\newblock U-net: Convolutional networks for biomedical image segmentation.
\newblock In \emph{International Conference on medical image computing and
  computer-assisted intervention}, 2015.

\bibitem[Russakovsky et~al.(2015)Russakovsky, Deng, Su, Krause, Satheesh, Ma,
  Huang, Karpathy, Khosla, Bernstein, Berg, and Fei-Fei]{russakovsky-imagenet}
O.~Russakovsky, J.~Deng, H.~Su, J.~Krause, S.~Satheesh, S.~Ma, Z.~Huang,
  A.~Karpathy, A.~Khosla, M.~Bernstein, A.~Berg, and L.~Fei-Fei.
\newblock Imagenet large scale visual recognition challenge.
\newblock \emph{International Journal of Computer Vision}, 2015.

\bibitem[Xia et~al.(2015)Xia, Cao, Wen, Hua, and
  Sun]{xia-unsupervised-discriminative-2015}
Y.~Xia, X.~Cao, F.~Wen, G.~Hua, and J.~Sun.
\newblock Learning discriminative reconstructions for unsupervised outlier
  removal.
\newblock In \emph{International Conference on Computer Vision}, 2015.

\end{thebibliography}

\appendix

\section{Dataset Details}\label{apd:data}

We attempted to obtain training samples from a diverse subset of the real-world distribution of vascular and emergency ultrasound imaging. In total, we included 5 datasets in this study, with each dataset containing multiple video sequences. Data was acquired from a phantom, a preexisting de-identified human subject \citep{mathai-uhfus-2019}, and a live-pig subject. The phantom that was used in this work was the CAE Blue Phantom femoral vascular access lower torso ultrasound training model (BPF1500-HP). In the human subjects, the arteries and veins in the palmar arch of the hand were imaged.

\noindent
\textbf{Scanners}: Three different ultrasound scanners were used for imaging the phantom, pig, and human subjects: (1) Fukuda Denshi portable (i.e. Point of Care Ultrasound, \emph{POCUS}) scanner with a 5-12 MHz transducer, (2) Diasus High-Frequency Ultrasound (\emph{HFUS}) scanner (Dynamic Imaging, UK) with a 10-22 MHz transducer, and (3) a VisualSonics Vevo 2100 Ultra High Frequency Ultrasound (\emph{UHFUS}) scanner with a 50MHz transducer. The images include diverse scan parameters and settings (e.g. gain values) and anatomical variations. Datasets representing this diversity of imaging settings include: \emph{human-single50}, from the UHFUS machine, \emph{phantom1-multi22}, from the HFUS machine, and \emph{phantom1-multi12}, from the lower frequency Fukuda Denshi machine. The numerical suffix for each dataset name represents the ultrasound frequency with which it was collected.

\noindent
\textbf{Phantom-Based Categories}: Three categories of imaged sequences were acquired from the phantom: (1) \emph{phantom1-multi12}, (2) \emph{phantom1-multi22}, and (3) \emph{phantom2-multi12}. The prefix ``phantom1'' represents image sequences collected from the left side of the phantom. On the other hand, the prefix ``phantom2'' represents data collected from the right side of the phantom, which contained different anatomy (muscles, liver, etc.) and other artifacts unknown to the model, resulting in a more complex dataset. Each of these datasets consisted of the following classes: arteries, veins, ligaments, and nerves.

\noindent
\textbf{Human Subject-Based Category}: The arteries and veins in the palmar arch of the hand \citep{mathai-fast-vessel-segmentation-2018} were imaged using the UHFUS (50 MHz) scanner, and a wide range of gain values were used (40-70 dB). Each sequence consisted of only one artery or vein, and the same expert labeled the vessel in each frame as an artery.  We refer to this UHFUS human category as \emph{human-single50}.

\noindent
\textbf{Live Pig-Based Category}: The femoral arteries and veins of a live pig were imaged using the Fukuda Denshi portable scanner with a 5-12 MHz transducer. We refer to this category as \emph{pig-multi12}.

\noindent
\textbf{Substantial Noise and Artifacts:} 
Of the 5 datasets, \emph{human-single50} consists of the most amount of noise and speckle as a result of its high frequency, then closely followed by \emph{phantom1-multi22} for the same reason. \emph{phantom2-multi12}, on the other hand, contains more artifacts as a result of the different positioning on the phantom.

\noindent
\textbf{Data Quantity:} The \emph{phantom1-multi12} category consisted of $12$ sequences, each containing $50$ frames totalling $600$ frames. The \emph{phantom1-multi22} category consisted of $18$ sequences, each containing $31$ frames totalling $558$ frames. The \emph{phantom2-multi12} category consisted of $3$ sequences, with each sequence containing $50$ frames totalling $150$ frames. For the \emph{human-single50} category, $10$ sequences were obtained in total from the left and right hands of 4 subjects \citep{mathai-uhfus-2019}, with each sequence containing $50$ frames totalling $500$ frames. Lastly, the \emph{pig-multi12} category consisted of $3$ sequences, with each sequence containing $40$ frames totalling $120$ frames. All classes (arteries, veins, ligaments and nerves) were present in (at least some frames of) every sequence in \emph{phantom1-multi-12}, \emph{phantom1-multi22}, and \emph{phantom2-multi12}. Only a single vessel class (of either artery or vein) was present in \emph{human-single50} whereas both artery and vein classes were present in \emph{pig-multi12}. Each frame was annotated by a single expert.

\section{Training Details}\label{apd:training}

For each of the encoder and decoder transfer learning scenarios, we set batch normalization to use the overall training data's statistics, as opposed to batch statistics, as that is often what is used in practice.

Each of the training datasets consisted of close to 600 training images each. Each of the datasets, including \emph{pig-multi12 (p12)} and \emph{phantom2-multi12 (ph2-12)}, contains 150 images for testing. Similar to \citep{mathai-uhfus-2019}, we trained the multi-class segmentation models by resizing each ultrasound B-scan to 256x256 pixels. Traditional (spatial) data augmentations were done by random flipping, rotating, blurring, and translating the training set, such that each experimental run's training set was increased to $\sim$ 12,000 images. All experiments were conducted using TensorFlow \citep{martin-tensorflow-2016}, training with the Adam optimizer \citep{kingma-adam-2014} on cross-entropy loss with a batch size of 16, learning rate of 0.0001 for pre-training, and learning rate of 0.000001 for fine-tuning. Final pixel-level probabilities were classified using softmax, and the results were evaluated using the Dice Similarity Coefficient (DSC).

\section{Results Details}\label{apd:results}

\begin{table*}[!t]
\centering\fontsize{9}{12}\selectfont 
\setlength\aboverulesep{0pt}\setlength\belowrulesep{0pt} 
\setlength{\tabcolsep}{7pt} 
\caption{Transfer Learning Experimental Results using Dice Coefficient metric (averaged over 3 runs) comparing across changes in anatomy and imaging settings. Encoder 1 (e-1) means that only the first block in the encoder is used for fine-tuning, encoder 2 (e-2) means 2 blocks in the encoder are used, decoder 4 (d-4) refers to only the first block on the decoder side, and so on. \emph{OOTD} is the arithmetic average of the Dice metric on the unseen datasets.}
\begin{adjustbox}{max width=\textwidth}
\begin{tabular}{cccccccccccc}
\toprule
{No.} & {Model} & {\emph{pt-data}} & {\emph{ft-data}} & {Method} & {\emph{h50}} & {\emph{ph1-12}} & {\emph{ph1-22}} & {\emph{ph2-12}} & {\emph{p12}} & {\emph{OOTD}} \\
\midrule
    1 & 1 & ph1-12 & h50 & e-1 & .499 $\pm$ .0003 & --- & .578 $\pm$ .0010 & .610 $\pm$ .0030 & .531 $\pm$ .0023 & .583 $\pm$ .0206 \\
    1 & 2 & ph1-12 & h50 & e-2 & .848 $\pm$ .0013 & --- & .614 $\pm$ .0001 & .592 $\pm$ .0113 & .596 $\pm$ .0001 & .600 $\pm$ .0116 \\
    1 & 3 & ph1-12 & h50 & e-3 & .814 $\pm$ .0010 & --- & .615 $\pm$ .0001 & .577 $\pm$ .0002 & .595 $\pm$ .0001 & .596 $\pm$ .0150 \\ 
    1 & 4 & ph1-12 & h50 & e-4 & .913 $\pm$ .0164 & --- & .614 $\pm$ .0001 & .596 $\pm$ .0051 & .597 $\pm$ .0009 & .602 $\pm$ .0086 \\
    1 & 5 & ph1-12 & h50 & e-5 & .906 $\pm$ .0001 & --- & .614 $\pm$ .0001 & .582 $\pm$ .0001 & .609 $\pm$ .0005 & .602 $\pm$ .0128 \\
    1 & 6 & ph1-12 & h50 & d-4 & .520 $\pm$ .0307 & --- & .383 $\pm$ .0333 & .626 $\pm$ .0083 & .275 $\pm$ .0175 & .428 $\pm$ .1485 \\ 
    1 & 7 & ph1-12 & h50 & d-3 & .759 $\pm$ .0016 & --- & .582 $\pm$ .0021 & .687 $\pm$ .0032 & .401 $\pm$ .0065 & .557 $\pm$ .1184 \\
    1 & 8 & ph1-12 & h50 & d-2 & .723 $\pm$ .0036 & --- & .599 $\pm$ .0031 & .690 $\pm$ .0020 & .392 $\pm$ .0063 & .560 $\pm$ .1247 \\
    1 & 9 & ph1-12 & h50 & d-1 & .959 $\pm$ .0009 & --- & .597 $\pm$ .0014 & .719 $\pm$ .0014 & .651 $\pm$ .0016 & .656 $\pm$ .0498 \\
    1 & 10 & ph1-12 & h50 & Full & .957 $\pm$ .0010 & --- & .546 $\pm$ .0001 & .657 $\pm$ .0001 & .632 $\pm$ .0003 & .612 $\pm$ .0383 \\
    \midrule
    2 & 11 & h50 & ph1-12 & e-1 & --- & .589 $\pm$ .0033 & .536 $\pm$ .0001 & .625 $\pm$ .0010 & .596 $\pm$ .0001 & .585 $\pm$ .0372 \\
    2 & 12 & h50 & ph1-12 & e-2 & --- & .491 $\pm$ .0003 & .545 $\pm$ .0001 & .577 $\pm$ .0002 & .623 $\pm$ .0001 & .582 $\pm$ .0319 \\
    2 & 13 & h50 & ph1-12 & e-3 & --- & .608 $\pm$ .0001 & .552 $\pm$ .0016 & .653 $\pm$ .0005 & .600 $\pm$ .0004 & .602 $\pm$ .0408 \\ 
    2 & 14 & h50 & ph1-12 & e-4 & --- & .629 $\pm$ 0001 & .604 $\pm$ .0002 & .667 $\pm$ .0001 & .595 $\pm$ .0002 & .622 $\pm$ .0317 \\ 
    2 & 15 & h50 & ph1-12 & e-5 & --- & .656 $\pm$ .0006 & .605 $\pm$ .0034 & .673 $\pm$ .0001 & .596 $\pm$ .0001 & .625 $\pm$ .0346 \\ 
    2 & 16 & h50 & ph1-12 & d-4 & --- & .483 $\pm$ .0003 & .497 $\pm$ .0001 & .564 $\pm$ .0002 & .629 $\pm$ .0001 & .563 $\pm$ .0540 \\ 
    2 & 17 & h50 & ph1-12 & d-3 & --- & .507 $\pm$ .0001 & .494 $\pm$ .0002 & .571 $\pm$ .0015 & .611 $\pm$ .0002 & .559 $\pm$ .0484 \\ 
    2 & 18 & h50 & ph1-12 & d-2 & --- & .508 $\pm$ .0003 & .491 $\pm$ .0014 & .557 $\pm$ .0005 & .601 $\pm$ .0001 & .550 $\pm$ .0452 \\ 
    2 & 19 & h50 & ph1-12 & d-1 & --- & .845 $\pm$ .0113 & .482 $\pm$ .0004 & .554 $\pm$ .0024 & .267 $\pm$ .0003 & .435 $\pm$ .1217 \\ 
    2 & 20 & h50 & ph1-12 & Full & --- & .919 $\pm$ .0002 & .366 $\pm$ .0017 & .417 $\pm$ .0041 & .201 $\pm$ .0011 & .328 $\pm$ .0919 \\
\bottomrule
\end{tabular}
\end{adjustbox}
\label{table:transfer-results-details-1}
\end{table*}

\begin{table*}[!t]
\centering\fontsize{9}{12}\selectfont 
\setlength\aboverulesep{0pt}\setlength\belowrulesep{0pt} 
\setlength{\tabcolsep}{7pt} 
\caption{Transfer Learning Experimental Results using Dice Coefficient metric (averaged over 3 runs) comparing across imaging settings. Encoder 1 (e-1) means that only the first block in the encoder is used for fine-tuning, encoder 2 (e-2) means 2 blocks in the encoder are used, decoder 4 (d-4) refers to only the first block on the decoder side, and so on. \emph{OOTD} is the arithmetic average of the Dice metric on the unseen datasets.}
\begin{adjustbox}{max width=\textwidth}
\begin{tabular}{cccccccccccc}
\toprule
{No.} & {Model} & {\emph{pt-data}} & {\emph{ft-data}} & {Method} & {\emph{h50}} & {\emph{ph1-12}} & {\emph{ph1-22}} & {\emph{ph2-12}} & {\emph{p12}} & {\emph{OOTD}} \\
\midrule
    3 & 21 & ph1-12 & ph1-22 & e-1 & .456 $\pm$ .0032 & --- & .561 $\pm$ .561 & .627 $\pm$ .0004 & .325 $\pm$ .0016 & .469 $\pm$ .1239 \\
    3 & 22 & ph1-12 & ph1-22 & e-2 & .769 $\pm$ .0013 & --- & .612 $\pm$ .0015 & .662 $\pm$ .0019 & .419 $\pm$ .0016 & .617 $\pm$ .1467 \\
    3 & 23 & ph1-12 & ph1-22 & e-3 & .794 $\pm$ .0003 & --- & .773 $\pm$ .0009 & .647 $\pm$ .0005 & .410 $\pm$ .0050 & .617 $\pm$ .1582 \\ 
    3 & 24 & ph1-12 & ph1-22 & e-4 & .700 $\pm$ .0025 & --- & .838 $\pm$ .0032 & .594 $\pm$ .0168 & .595 $\pm$ .0040 & .630 $\pm$ .0511 \\
    3 & 25 & ph1-12 & ph1-22 & e-5 & .729 $\pm$ .0042 & --- & .852 $\pm$ .0040 & .601 $\pm$ .0059 & .531 $\pm$ .0064 & .620 $\pm$ .0820 \\
    3 & 26 & ph1-12 & ph1-22 & d-4 & .179 $\pm$ .0007 & --- & .753 $\pm$ .0045 & .375 $\pm$ .0004 & .200 $\pm$ .0008 & .251 $\pm$ .0878 \\ 
    3 & 27 & ph1-12 & ph1-22 & d-3 & .201 $\pm$ .0001 & --- & .840 $\pm$ .0029 & .403 $\pm$ .0006 & .218 $\pm$ .0001 & .274 $\pm$ .0916 \\
    3 & 28 & ph1-12 & ph1-22 & d-2 & .218 $\pm$ .0004 & --- & .849 $\pm$ .0058 & .419 $\pm$ .0038 & .238 $\pm$ .0002 & .292 $\pm$ .0904 \\
    3 & 29 & ph1-12 & ph1-22 & d-1 & .256 $\pm$ .0015 & --- & .851 $\pm$ .0029 & .464 $\pm$ .0006 & .221 $\pm$ .0002 & .314 $\pm$ .1075 \\
    3 & 30 & ph1-12 & ph1-22 & Full & .481 $\pm$ .0151 & --- & .923 $\pm$ .0001 & .471 $\pm$ .0025 & .227 $\pm$ .0009 & .393 $\pm$ .1175 \\
    \midrule
    4 & 31 & ph1-22 & ph1-12 & e-1 & .597 $\pm$ .0004 & .684 $\pm$ .0002 & --- & .539 $\pm$ .0018 & .375 $\pm$ .0023 & .504 $\pm$ .0943 \\
    4 & 32 & ph1-22 & ph1-12 & e-2 & .664 $\pm$ .0005 & .708 $\pm$ .0006 & --- & .592 $\pm$ .0515 & .344 $\pm$ .0001 & .533 $\pm$ .1541 \\
    4 & 33 & ph1-22 & ph1-12 & e-3 & .650 $\pm$ .0042 & .823 $\pm$ .0002 & --- & .595 $\pm$ .0026 & .315 $\pm$ .0015 & .520 $\pm$ .1466 \\
    4 & 34 & ph1-22 & ph1-12 & e-4 & .545 $\pm$ .0009 & .936 $\pm$ .0034 & --- & .642 $\pm$ .0013 & .326 $\pm$ .0024 & .505 $\pm$ .1605 \\
    4 & 35 & ph1-22 & ph1-12 & e-5 & .555 $\pm$ .0071 & .969 $\pm$ .0001 & --- & .732 $\pm$ .0019 & .338 $\pm$ .0047 & .541 $\pm$ .1702 \\
    4 & 36 & ph1-22 & ph1-12 & d-4 & .193 $\pm$ .0001 & .740 $\pm$ .0038 & --- & .579 $\pm$ .0002 & .234 $\pm$ .0002 & .335 $\pm$ .1729 \\
    4 & 37 & ph1-22 & ph1-12 & d-3 & .233 $\pm$ .0005 & .816 $\pm$ .0019 & --- & .498 $\pm$ .0017 &  .204 $\pm$ .0005 & .312 $\pm$ .1320 \\
    4 & 38 & ph1-22 & ph1-12 & d-2 & .269 $\pm$ .0005 & .826 $\pm$ .0002 & --- & .604 $\pm$ .0011 & .216 $\pm$ .0001 & .363 $\pm$ .1716 \\
    4 & 39 & ph1-22 & ph1-12 & d-1 & .329 $\pm$ .0002 & .965 $\pm$ .0001 & --- & .543 $\pm$ .0022 & .241 $\pm$ .0011 & .371 $\pm$ .1275 \\
    4 & 40 & ph1-22 & ph1-12 & Full & .370 $\pm$ .0024 & .958 $\pm$ .0001 & --- & .634 $\pm$ .0002 & .203 $\pm$ .0009 & .403 $\pm$ .1776 \\
\bottomrule
\end{tabular}
\end{adjustbox}
\label{table:transfer-results-details-2}
\end{table*}

\subsection{Encoder vs. Decoder \emph{OOTD} Paired-Difference Statistical Significance Testing}
To ensure that similar feature representational power is represented across a pair in our statistical hypothesis tests, we paired matching numbers of blocks, e.g. ``encoder 1'' with ``decoder 4,'' ``encoder 2'' (i.e. encoder blocks 1-2) with ``decoder 3'' (i.e. decoder blocks 4-3), and so on. To ensure equal sample sizes, we ignore encoder block 5, the bottleneck block of the U-Net architecture. $H_0: \mu_{encoder} = \mu_{decoder}$. $H_1: \mu_{encoder} \geq \mu_{decoder}$. To evaluate for statistical significance, we use the Wilcoxon Test to compare the differences in the \emph{OOTD} scores between the matching contiguous-block pairs for each experiment. We use a significance level of $\alpha = 0.05$. We calculated the following p-values for experiments 1, 2, 3, and 4, respectively: .1182, .0010, .0010, .0011. Fine-tuning the encoder subsequences resulted in a greater out-of-domain generalization performance on all experiments except for experiment 1. We note that this may be due to the greater number of classes in the \emph{pt-data} enabling the decoder to learn more generalized feature representations, although more experimentation will be needed.

\subsection{Contiguous Encoder/Decoder Blocks vs. \emph{OOTD} Scores Linear Relationship Statistical Significance Testing}
To evaluate the statistical significance of the linear relationship between an increasing number of encoder blocks and the \emph{OOTD} scores, we use the coefficient p-value after fitting an ordinary least squares (OLS) regression line to each of the experiments' samples. For example, for experiment 1, we would first fit an OLS line to the averaged \emph{OOTD} scores in Table \ref{table:transfer-results-details-1} and then test using the coefficient p-value. We found statistically significant linear relationships across all experiments except for the encoders of experiment 4, which had a relatively constant relationship. Experiment 2 on the decoder side was a statistically significant negative linear relationship, which is discussed in Section \ref{sec:experiments-results}. Tables \ref{hypothesis-testing-encoder-ootd} and \ref{hypothesis-testing-decoder-ootd} show our results. $H_0: \beta_1 = 0$. $H_1: \beta_1 \neq 0$.

\subsection{Autoencoder Reconstruction Experiment Details}
It was noted in \citep{xia-unsupervised-discriminative-2015} that statistically similar data produced lower autoencoder reconstruction errors; data with additional noise/outliers often resulted in higher errors. Using that observation, we train a convolutional autoencoder model on each of the datasets separately and note down the average training loss to quantify the difficulty of each of the three training datasets, \emph{h50}, \emph{ph1-12}, and \emph{ph1-22}. Our convolutional autoencoder consists of 3 convolution-max-pool blocks on the encoder side and 3 convolution-upsampling blocks on the decoder side. Each convolution layer had 3x3-dimension kernels followed by ReLU activation. The final layer consisted of 1 channel, for each of the ultrasound images. The convolutional autoencoder was trained for 5 epochs with a batch size of 16, learning rate of .001, and cross entropy loss. We trained the autoencoder on each dataset 5 times and noted down the reconstruction errors as follows: \emph{h50} (11.379 $\pm$ .002), \emph{ph1-12} (12.347 $\pm$ .001), and \emph{ph1-22} (12.782 $\pm$ .001). We further performed Wilcoxon statistical significance tests for the following cases: (1) $H_0: \mu_{h50} = \mu_{ph1-12}$. $H_1: \mu_{h50} \leq \mu_{ph1-12}$. (2) $H_0: \mu_{ph1-12} = \mu_{ph1-22}$. $H_1: \mu_{ph1-12} \leq \mu_{ph1-22}$ and calculated the p-values: .0253 and .0258, respectively. This follows along with our observations noted in Section \ref{sec:experiments-results}; because the \emph{ft-data} was more challenging, the model may have needed additional blocks for a better latent representation. Related patent pending.

\begin{table}[h]
    \caption[]{Encoder-side vs. \emph{OOTD} Scores Linearity Statistical Significance Results. Statistical significance at $\alpha$ = 0.05 bolded.}
    \label{hypothesis-testing-encoder-ootd}
    \begin{tabular}{|c||c||c|}
    \hline
    No. & P-Value & 95\% Confidence Interval \\
    \hline
    \hline
    1 & \textbf{.008} & (.001, .007) \\
    \hline
    2 & \textbf{.001} & (.009, .015) \\
    \hline
    3 & \textbf{.017} & (.008, .055) \\
    \hline
    4 & .158 & (-.002, .013) \\
    \hline
    \end{tabular}
\end{table}

\begin{table}[h]
    \caption[]{Decoder-side vs. \emph{OOTD} Linearity Results. Note that experiment 2 has a negative linear relationship.}
    \label{hypothesis-testing-decoder-ootd}
    \begin{tabular}{|c||c||c|}
    \hline
    No. & P-Value & 95\% Confidence Interval \\
    \hline
    \hline
    1 & \textbf{.001} & (.050, .092) \\
    \hline
    2 & \textbf{.011} & (-.066, -.013) \\
    \hline
    3 & \textbf{.001} & (.020, .022) \\
    \hline
    4 & \textbf{.007} & (.007, .031) \\
    \hline
    \end{tabular}
\end{table}

\end{document}